\documentclass[conference]{IEEEtran}
\IEEEoverridecommandlockouts
\usepackage{cite}
\usepackage{amsmath,amssymb,amsfonts}
\usepackage{algorithm}
\usepackage{algpseudocode}
\usepackage{graphicx}
\usepackage{textcomp}
\usepackage{xcolor}
\usepackage{color}
\usepackage{tikz}
\usepackage{xspace}
\usepackage{subcaption}
\usepackage{verbatim}
\usepackage{float}
\usepackage{changepage}

\colorlet{pink1}{red!40}
\colorlet{blue1}{cyan!60}
\colorlet{green1}{green!40}

\newlength{\twosubht}
\newsavebox{\twosubbox}

\def\BibTeX{{\rm B\kern-.05em{\sc i\kern-.025em b}\kern-.08em
    T\kern-.1667em\lower.7ex\hbox{E}\kern-.125emX}}
\begin{document}

\title{Gaussian path model library for intuitive robot motion programming by demonstration\\}

\author{Samuli Soutukorva$^{1}$, Markku Suomalainen$^{1}$, Martin Kollingbaum$^{2}$ and Tapio Heikkilä$^{1}$%
\thanks{$^{1}$~VTT Technical Research Centre of Finland Ltd,
Oulu, Finland
        {\tt\small firstname.lastname@vtt.fi}}%
\thanks{$^{2}$~
        {\tt\small  mjkollingbaum@hotmail.com}}%
}

\maketitle

\begin{abstract}
This paper presents a system for generating Gaussian path models from teaching data representing the path shape. In addition, methods for using these path models to classify human demonstrations of paths are introduced. By generating a library of multiple Gaussian path models of various shapes, human demonstrations can be used for intuitive robot motion programming. A method for modifying existing Gaussian path models by demonstration through geometric analysis is also presented.
\end{abstract}


\section{Introduction}
 
Programming by Demonstration (PbD) is one of the commonly proposed solutions for intuitive robot programming, where a user \textit{shows} the desired trajectory, or skill, to a robot \cite{ravichandar_survey}. PbD, thus, has the potential to enable domain experts with limited expertise in robotics to teach a robot what actions to perform, purely through demonstrations, allowing a wider use of robotics across multiple industries. In general, a multitude of demonstrations is required to provide the amount of data for robotic systems to properly learn a demonstrated trajectory. Domain experts, however, may prefer robotic systems where one demonstration is sufficient for the robot to deploy a suitable skill that properly imitates the demonstrated trajectory.

In this paper, an approach for programming robot paths with a single demonstration is presented that utilises pre-trained models of prototypical movements a robot may perform (Gaussian path models) and methods of adapting and parameterising such models according to a single demonstration by a domain expert, in order to generate the required robot skill. Gaussian path models are general-purpose prototypical movements (such as, for example, half-circle and other path segments, see Fig. \ref{fig:path_library}) and are generated themselves in a pre-training step, where the training data is derived from either demonstrations or synthetically produced. A library of these path models can then be used in a programming-by-demonstration effort, where a single demonstration provided by a domain expert is used to retrieve the best-fit path model from such a library and parameterize this model with respect to the demonstration.

The contribution of this paper is a set of algorithms for building such a library of Gaussian path models, as well as for matching demonstrations to path models and adapting them for a specific programming purpose. It is also shown how path models can be efficiently modified by a domain expert to create the required path data for programming a robot, without the need to train a new path model.

\section{Related work and background}
\label{sec:related}

There is a wide variety of ways to extract suitable data from human demonstrations to generate a condensed representation that attempts to capture the demonstrated path, or trajectory. Splines are widely used in robotics in general, as well as in PbD \cite{aleotti}. 
Lately, different kinds of Gaussian-based models have proven popular; a good overview with examples is provided in \cite{calinon_2016}. Of the presented models, for example, Hidden Markov Models (HMM) \cite{sugiura} 
and Gaussian mixture model-Gaussian Mixture Regression (GMM-GMR) have been used for motion captured data \cite{sun}. Even Dynamic Movement Primitives (DMP) have been created from GMM \cite{chen}. 
Further Gaussian models are, for example, Infinite Gaussian Mixture Models (IGMM) \cite{kruger}. 
Several proposals exist to generate libraries such as in this paper. In \cite{meier}, a dynamic movement primitive (DMP) library is used to segment trajectories consisting of sequences of movement primitives. Methods for segmenting human demonstrations into movement primitives and compiled to movement primitive libraries have been proposed \cite{lioutikov2015, lioutikov2017}. Position and force-torque data from human demonstrations have been utilized to recognize predefined skills from a skill library \cite{eiband2023,eiband2021}.
A common theme to all the mentioned Gaussian-based methods is that a suitable number of keypoints characterizing a path is required to preserve the shape of the movement. However, details regarding the identification of keypoints is often neglected in these papers. For time-variant demonstrations, methods such as Non-Maximum Suppression algorithm (NMS) has been proposed \cite{song}. In contrast, the proposed method uses a Gaussian Hidden Markov Model (GHMM) based approach, and presents practical methods for managing the number of keypoints. Moreover, we present a way for on-the-fly modifications, a method proposed before, however, with different representations and without focusing on the keypoints \cite{aleotti, ginesi}.

\section{Methods}
\label{sec:methods}

In the approach presented in this paper, a Gaussian path model is created from multiple teaching data sets of path data. These teaching data sets are first brought into a canonicalized form (\ref{intro:canon}) and then reduced to an approximation of the original demonstration (\ref{intro:decim}) as sequences of path keypoints. With this data, a Gaussian model of the path is created from the means and covariances of the path keypoints (\ref{intro:gausmodel}). Such a Gaussian model can then be used to classify demonstrations (\ref{intro:pathrecog}) and utilized in intuitive robot programming. 
This approach shares similarities with the construction of GHMM, however, it is limited to linear sequences only.

\subsection{Canonicalization}\label{intro:canon}
In order for a path model to become flexibly utilizable for arbitrary path recognition and robot programming, the path model must be agnostic with regards to scale, position, and the orientation of the path data. This is achieved through storing the path model in a canonicalized form of the path, where the only feature conserved from the teaching data is the shape of the path (Algorithm 1). The canonicalization of the path heavily relies on Principal Component Analysis (PCA).
The rotation matrix for re-orienting the path is defined by the eigenvectors of the sample sets of sample data points (Algorithm 1, line 2). To ensure a right-hand coordinate system defined by the eigenvectors, the rotation matrix is checked and adjusted by changing the sign of the third eigenvector (i.e., the Z-axis), if needed. 
First, the path point data is centralized around the mean of the data set (line 1 in Algorithm 1). Second, the path points are scaled (= normalized) by the standard deviations of the data set as the square roots of the eigenvalues of the data (line 10 in Algorithm 1). The normalized scale is then defined as the inverse of the square root of the dot product of the square roots of the eigenvalues. This equals to the sum of the vectors defined by the eigenvalues.
After the centralization and scaling, the data is rotated with the rotation matrix. 

\begin{algorithm}
\caption{Canonicalization of 3-D path}\label{alg:canonicalization}
\hspace*{\algorithmicindent} \textbf{Input:} $n$ data sets of 3-D points \\
\hspace*{\algorithmicindent} \textbf{Output:} Canonicalized 3-D points
\begin{algorithmic}[1]
    \State{Centralize data}
    \State{Rotation matrix $\leftarrow$\ PCA(centralized data).eigenvectors} \label{line:PCA1}
    \If{left handed coordinate frame}
        \State{z-axis = -z-axis} \Comment{Left-handed $\rightarrow$ right-handed}
    \EndIf
    \For {each set in data sets} \label{line:canonforset}
        \State{Centralize the set}
        \State{Calculate eigenvalues of the set}\label{line:PCA2}
        \State{$\sigma_x, \sigma_y, \sigma_z = \sqrt{\text{eigenvalues}}$}
        \State{$\text{scale} = \frac{1}{\sqrt{[\sigma_x, \sigma_y, \sigma_z] \cdot [\sigma_x, \sigma_y, \sigma_z]}}$}
        \State{Scale the centralized set}
        \State{Rotate the set (line \ref{line:PCA1}})
    \EndFor 
    \State{Return canonicalized 3-D points}
\end{algorithmic}
\end{algorithm}

\subsection{Decimation algorithms}\label{intro:decim}
As the teaching data for a path can consist of a large amount of points, the path data is decimated to find the set of keypoints as the least amount of data that is still representative of the demonstrated path. This approximation allows the path recognition to be done on a limited number of keypoints while conserving the shape of the path.
The approximation of the canonicalized path is generated by two polyline decimation algorithms. The Ramer–Douglas–Peucker algorithm (RDP) \cite{ramer} \cite{douglas} removes points by finding the point furthest from a generated line segment and comparing that distance to a given tolerance parameter. If the tolerance is exceeded, the point is kept and the process is repeated recursively. The Visvalingam–Whyatt algorithm (VW) \cite{visvalingam} removes points by generating triangles formed by adjacent points, finding the smallest triangle and comparing its area to a given tolerance parameter. If the tolerance is exceeded, the point is kept and the process is repeated. 

These decimation algorithms are used in sequence before generating a Gaussian model of a path from multiple sets of teaching data: RDP is used to find an appropriate level of decimation (i.e., the number of keypoints) for each generated path model. VW is then utilized to decimate the teaching data sets so that each of them contains an identical number of keypoints. This simplifies the generation of the Gaussian path model as the teaching data sets have the same number of keypoints, and also enables correct matching in path recognition. 

\subsection{Generating Gaussian path models}\label{intro:gausmodel}
\begin{figure*}
    \centering 
    \begin{subfigure}{0.32\textwidth}
        \includegraphics[width=\textwidth]{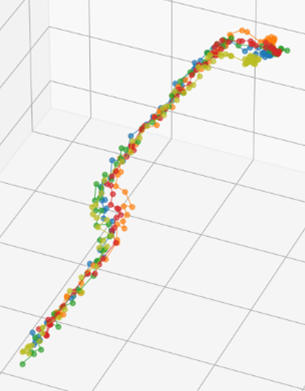}
        \caption{Canonicalized teaching data of the path.}
        \label{fig:teachdata}
    \end{subfigure}
    \hfill
    \begin{subfigure}{0.32\textwidth}
        \includegraphics[width=\textwidth]{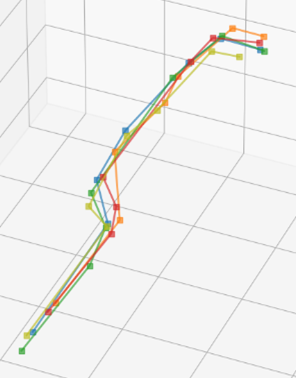}
        \caption{Keypoints from teaching data.}
        \label{fig:keypoints}
    \end{subfigure}
    \hfill
    \begin{subfigure}{0.327\textwidth}
        \includegraphics[width=\textwidth]{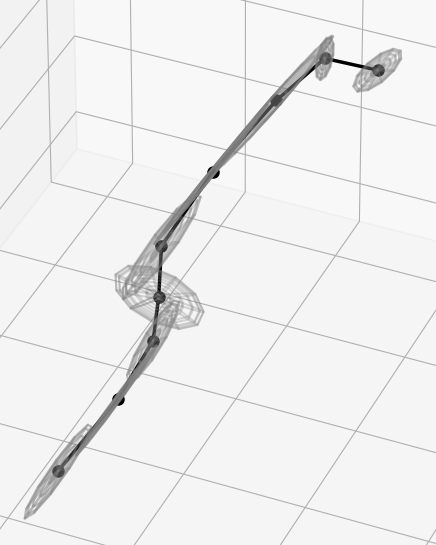}
        \caption{Gaussian model of the path.}
        \label{fig:gausmodel}
    \end{subfigure}     
    \caption{The sequence of creating a Gaussian path model from teaching data.}
    \label{fig:modelcreat}
\end{figure*}

Fig. \ref{fig:modelcreat} presents an example of creating a Gaussian path model from collected teaching data. From multiple sets of teaching data representing the same path (Fig. \ref{fig:teachdata}), a Gaussian model of the path is generated (Algorithm 2). First, the teaching data is canonicalized (Algorithm 1) and filtered with a Gaussian filter (Algorithm 2, line 2). Keypoints for the path model (Fig. \ref{fig:keypoints}) are then retrieved from the teaching data with the RDP and VW decimation algorithms (Algorithm 2, line 3). Mean values and covariances of the path keypoints are calculated from the path keypoints (Algorithm 2, lines 5-7). These mean values and covariances are then used to generate a Gaussian model of the path (Fig. \ref{fig:gausmodel}).

\begin{algorithm}
\caption{Create a Gaussian model from training data}\label{alg:gausmodel}
\hspace*{\algorithmicindent} \textbf{Input:} 3-D path points \\
\hspace*{\algorithmicindent} \textbf{Output:} Gaussian 3-D path model 
\begin{algorithmic}[1]
\State{Canonicalized 3-D path $\leftarrow$ Canonicalize(3-D path points)}
\Statex \Comment{Algorithm 1}
\State{Apply Gaussian filter to the canonicalized path data}
\State{Keypoint count $\leftarrow$ RDP(path data, epsilon)\label{line:RDP}}
\State{Path keypoint sets $\leftarrow$ VW(path data, keypoint count)\label{line:VW1}}
\For{keypoint in path keypoint sets}
    \State{Calculate the means for each keypoint:\label{line:means}}
    \Statex\hspace{\algorithmicindent}{$mean_i \leftarrow\ Mean(keypoint~set_1[keypoint_i],$}
    \Statex\hspace{\algorithmicindent}
            \hspace{\algorithmicindent}
            \hspace{\algorithmicindent}
            \hspace{\algorithmicindent}
            \hspace{\algorithmicindent}
            {$keypoint~set_2[keypoint_i],$}
    \Statex\hspace{\algorithmicindent}
            \hspace{\algorithmicindent}
            \hspace{\algorithmicindent}
            \hspace{\algorithmicindent}
            \hspace{\algorithmicindent}  
            {$...,$}
    \Statex\hspace{\algorithmicindent}
        \hspace{\algorithmicindent}
        \hspace{\algorithmicindent}
        \hspace{\algorithmicindent}
        \hspace{\algorithmicindent}     
        {$keypoint~set_{n-1}[keypoint_i],$}
    \Statex\hspace{\algorithmicindent}
        \hspace{\algorithmicindent}
        \hspace{\algorithmicindent}
        \hspace{\algorithmicindent}
        \hspace{\algorithmicindent}     
        {$keypoint~set_n[keypoint_i])$}
    \State{Calculate the covariance for each keypoint:}
    \Statex\hspace{\algorithmicindent}
        {$cov_i \leftarrow\ Covariance(keypoint~set_1[keypoint_i],$}
    \Statex\hspace{\algorithmicindent}
        \hspace{\algorithmicindent}
        \hspace{\algorithmicindent}
        \hspace{\algorithmicindent}
        \hspace{\algorithmicindent}   
        \hspace{\algorithmicindent}      
        {$keypoint~set_2[keypoint_i],$}
    \Statex\hspace{\algorithmicindent}
        \hspace{\algorithmicindent}
        \hspace{\algorithmicindent}
        \hspace{\algorithmicindent}
        \hspace{\algorithmicindent}   
        \hspace{\algorithmicindent}      
        {$...,$}
    \Statex\hspace{\algorithmicindent}
        \hspace{\algorithmicindent}
        \hspace{\algorithmicindent}
        \hspace{\algorithmicindent}
        \hspace{\algorithmicindent}   
        \hspace{\algorithmicindent}      
        {$keypoint~set_{n-1}[keypoint_i],$}
    \Statex\hspace{\algorithmicindent}
        \hspace{\algorithmicindent}
        \hspace{\algorithmicindent}
        \hspace{\algorithmicindent}
        \hspace{\algorithmicindent}   
        \hspace{\algorithmicindent}      
        {$keypoint~set_n[keypoint_i])$}
\EndFor
\State{Gaussian model of the path $\leftarrow$\ Mean values of keypoints, covariances of keypoints}
\State{Return Gaussian model of the path}
\end{algorithmic}
\end{algorithm}

For the Gaussian model to be representative of the original path and exploitable in path recognition, the teaching data has to meet some basic requirements. 
There has to be sufficient variability in the teaching data to create a path model that can be used in path recognition. With too little variability in the teaching path data, the paths cannot be classified as path models, as the path recognition relies on the covariances of the detected keypoints.

The teaching data has to be constructed of more than four separate sets of path points in order to produce variability in all three dimensions for each path model keypoint. 
For 3-D points, the covariance matrix associated with each keypoint has the rank of three if the number of sample sets is four or more. With only three sets of teaching data, each covariance associated with a Gaussian path model's keypoint would have the rank two. This would reduce the covariance matrix to conform to a plane (and with just two sets of teaching data, the covariance matrix would conform to a line).  

As the Gaussian path model is approximated from the original path (\ref{intro:decim}), a suitable level of decimation has to be defined. The performance of a Gaussian path model in path recognition varies as the level of decimation is changed. For example, with high enough decimation (i.e., very low keypoint count) the scores obtained in path recognition do not get very low with incorrect path models (as the path specific score is the sum of keypoint specific scores). This raises the possibility of an incorrect recognition.

\subsection{Path recognition}\label{intro:pathrecog}
The generated Gaussian path models can be utilized in recognizing paths demonstrated by a human and further used as paths for robot motion programming. Algorithm 3 outlines the procedure of recognizing a demonstrated path as a stored Gaussian path model. First, the demonstration of the path is canonicalized (Algorithm 1) and the canonicalized path data is then filtered with a Gaussian filter (Algorithm 3, line 2). 

 The canonicalized and filtered path demonstration is approximated with the VW decimation algorithm, using the candidate model's parameter for the number of keypoints (Algorithm 3, lines 4-5). This ensures that the decimated and canonicalized data of the demonstration has the same number of keypoints as the Gaussian model it is being compared to. The canonicalized path demonstration can now be compared to existing Gaussian path models (i.e., candidate models). In line 6 of Algorithm 3, the score for each keypoint in the demonstration is defined by calculating the loglikelihood ($\mathcal{L}$) for that keypoint ($x$), the corresponding keypoint of the candidate model ($\mu$), its covariance matrix ($\sum$) and the rank of the covariance matrix ($k$) (in our case $k$ equals to 3): 

\begin{equation}\label{eq:loglikelihood}
    \mathcal{L} = \log\frac{\exp(-\frac{1}{2}(x-\mu)^T\sum^{-1}(x-\mu))}{\sqrt{(2\pi)^k\det\sum}}
\end{equation}
 
These keypoint-specific scores are summed up to get the overall score for the path (Algorithm 3, line 8). By comparing the scores of multiple Gaussian path models, the best matching path model can be found for the demonstrated path.

\begin{algorithm}
\caption{Path recognition}\label{alg:pathrecog}
\hspace*{\algorithmicindent} \textbf{Input:} Single demonstration of a 3-D path \\
\hspace*{\algorithmicindent} \textbf{Output:} Result Gaussian model
\begin{algorithmic}[1]
\State{Canonical demonstration $\leftarrow$ Canonicalize(demonstration data set)}\Comment{Algorithm 1}
\State{Apply Gaussian filter to the canonicalized data set}
\For{each candidate model in model library}
    \State{Keypoint count $\leftarrow$ Candidate model's keypoint count \label{line:getkeyptcnt}}
    \State{Demonstration keypoints $\leftarrow$ VW(demonstration}
    \Statex{data set, keypoint count)\label{line:VW2}}
    \For{each keypoint in demonstration keypoints}
        \Statex{Keypoint score  $\leftarrow$ loglikelihood(demonstration keypoint, candidate model keypoint mean, candidate model keypoint covariance)\label{line:likelihood}}\Comment{(1)}
    \EndFor
    \State{$Model~score = \sum_{i=1}^{n} keypoint~score_i$}
\EndFor
\State{Best model score $\rightarrow$ result model}
\State{Return result Gaussian model}
\end{algorithmic}
\end{algorithm}

\subsection{Exploitation of Gaussian models and path recognition}\label{intro:decanon}
Once a demonstrated path is recognized through a specific path model, the keypoints of the matching Gaussian path model can be decanonicalized with Algorithm 4. Here, the once canonicalized path model is scaled, transformed, and rotated to the  configuration set by the demonstration data. This allows the Gaussian path model's keypoint means to be utilized in robot motion programming, fitted to the demonstration's scale, orientation, and position. Here, the demonstration of the path can be thought of as a localization action for the path keypoints that are fetched from the Gaussian path model through the path recognition procedure.

\begin{algorithm}
\caption{Decanonicalization of 3-D path}\label{alg:decanonicalization}
\hspace*{\algorithmicindent} \textbf{Input:} Canonicalized 3-D path, reference scale, reference rotation, reference translation \\
\hspace*{\algorithmicindent} \textbf{Output:} Decanonicalized 3-D path
\begin{algorithmic}[1]
    \For {each point in path} 
        \State{Revert orientation}
        \State{Revert scale}
        \State{Revert centralization}
    \EndFor 
\State{Return decanonicalized path}
\end{algorithmic}
\end{algorithm}

From multiple Gaussian path models, a path model library, consisting of a variety of paths, can be created and utilized in path recognition, and, furthermore, in robot motion programming by demonstration. In addition to different path shapes, variations of the same path can be generated from the same teaching data to bring versatility to the library. For example, the direction of the path in the model can be reversed allowing the demonstration to be in either direction along the path.
More critically, also flipped path models have to be created. As the direction of the eigenvectors (Algorithm 1, line 2) can be one of two, and which cannot be predicted, the orientation of the path can also vary in two different variations per eigenvector. For the path recognition (Algorithm 3) to work robustly, flipped variations for the Gaussian path model are generated for the library. Flipped variations have the rotation matrix of the canonicalized path model, but rotated 180\textdegree~around a coordinate axis.

\subsection{Correcting Gaussian path models}\label{intro:pathcorr}
As through the decimation step, the created Gaussian path models are based on a very limited number of keypoints, they can be adjusted with relative ease by manipulating the keypoint sequence of the Gaussian path model.
In a case, where a path has been demonstrated and recognized to be best represented by a particular Gaussian path model (i.e., the Gaussian path model's keypoints are close to those of a demonstration), the path model's keypoints can be revised by demonstration with a geometric analysis of the demonstrated correction keypoints and the Gaussian path model's keypoints (Algorithm 5). This geometric analysis finds the redundant keypoints from the path model and inserts the correction keypoints to replace them, forming a new path keypoint sequence. This allows for the creation of a new path utilizing the Gaussian path model's keypoints and the revised (or correction) keypoints without having to generate a new path model. 

\begin{figure*}[htpb]
\sbox\twosubbox{%
  \resizebox{\dimexpr.98\textwidth-1em}{!}{%
    \includegraphics[height=5cm]{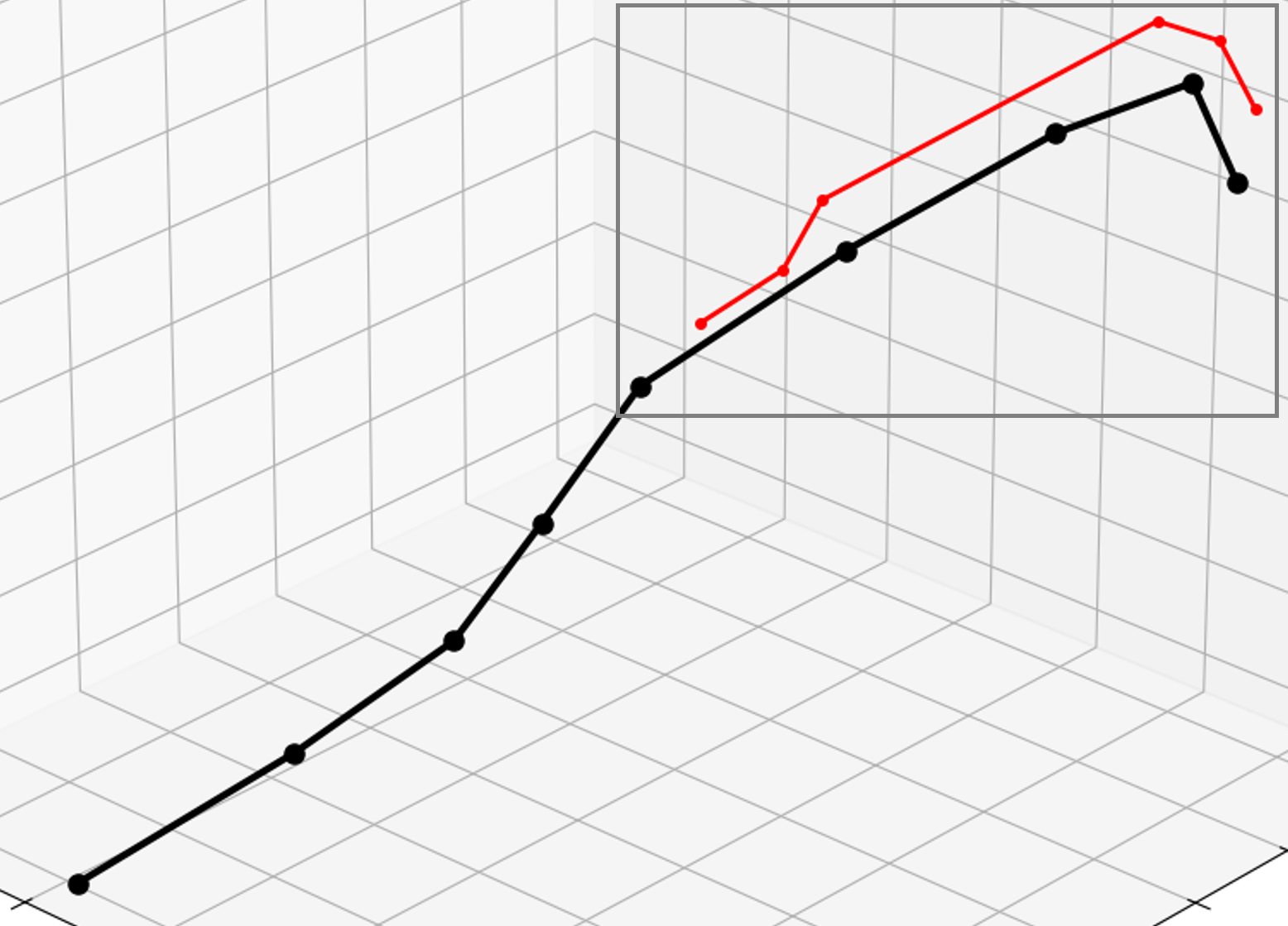}%
    \includegraphics[height=5cm]{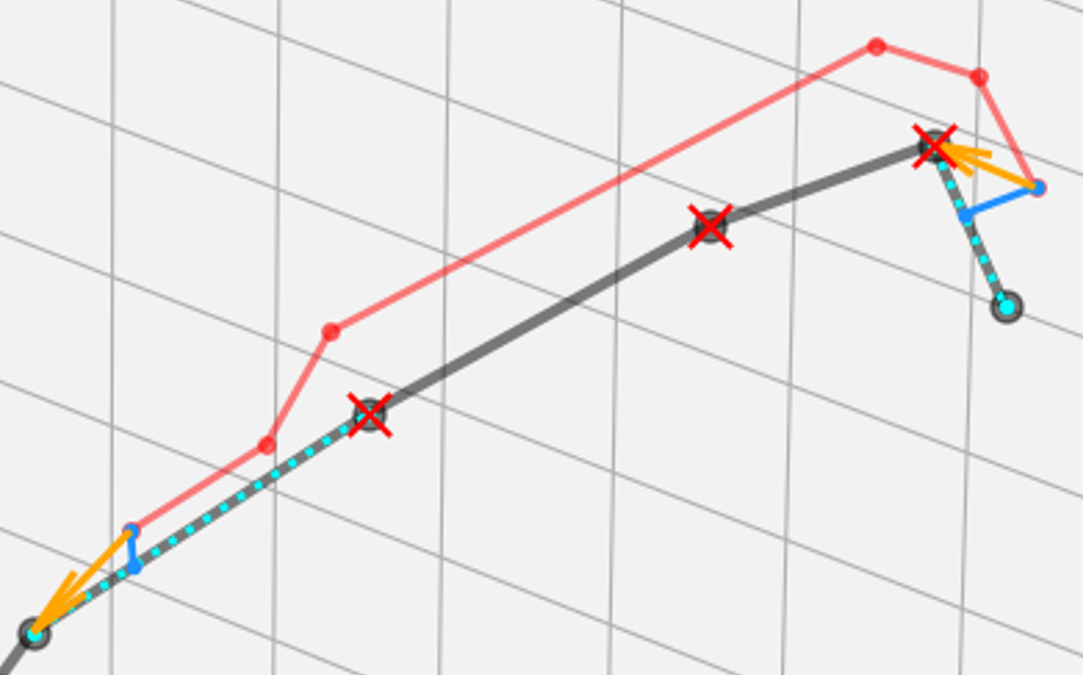}%
    \includegraphics[height=5cm]{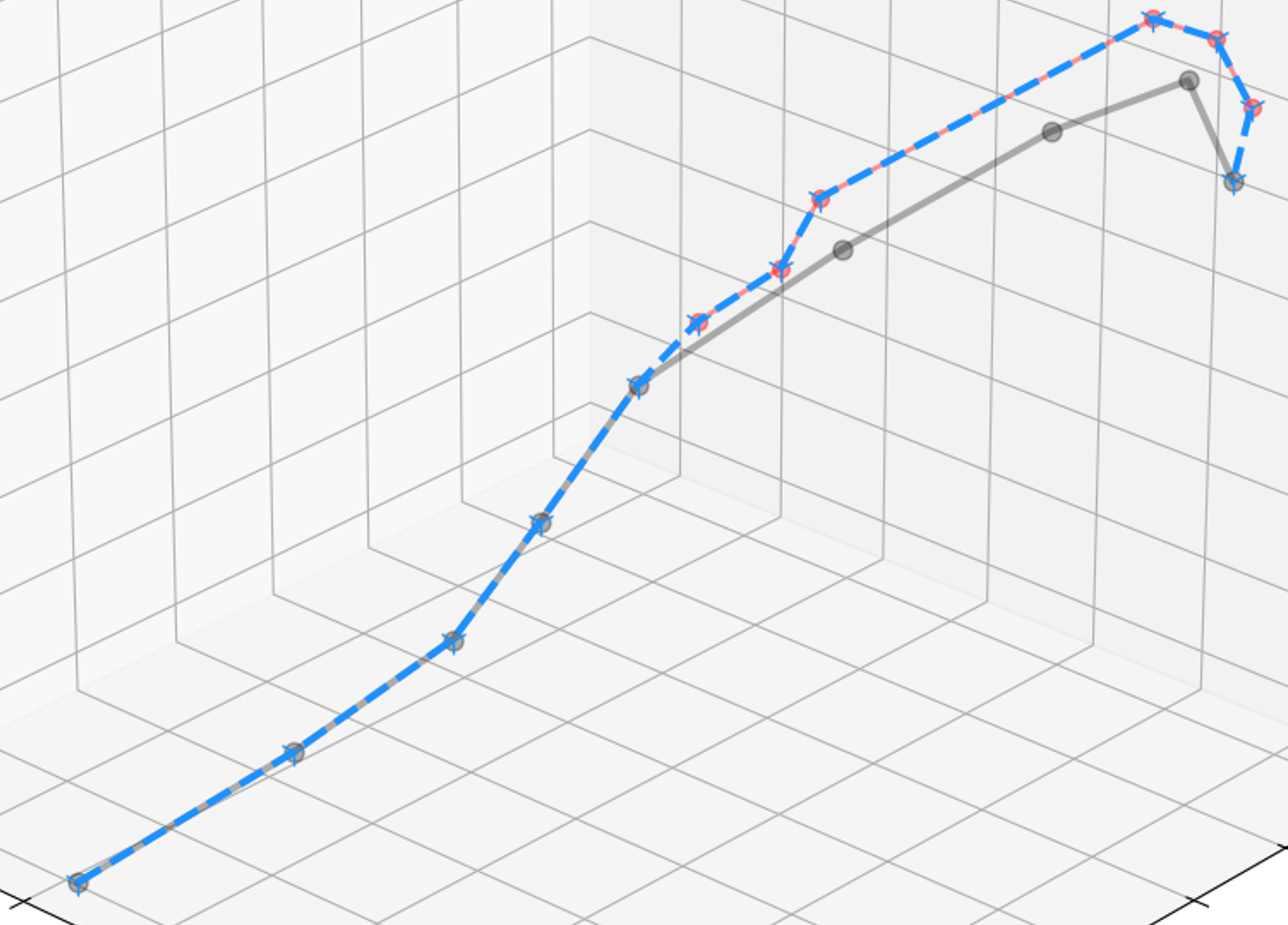}%
  }%
}
\setlength{\twosubht}{\ht\twosubbox}
\centering
\subcaptionbox{Path model keypoints (black) and correction keypoints (red).\label{fig:pathcorr_paths}}{%
  \includegraphics[height=\twosubht]{path_correction_paths_close_highlighted.png}%
}\quad
\subcaptionbox{Geometric analysis of path correction.\label{fig:pathcorr_analysis}}{%
  \includegraphics[height=\twosubht]{path_correction_analysis_zoom.png}%
}\quad
\subcaptionbox{Corrected path (blue).\label{fig:pathcorr_correctedpath}}{%
  \includegraphics[height=\twosubht]{path_correction_corrected_path_close.png}%
}
\caption{The sequence of the path correction.}\label{fig:pathcorr}
\end{figure*}

Fig. \ref{fig:pathcorr} displays the sequence of the path correction. First, a path must be demonstrated and recognized, and a correction to this path must be demonstrated and approximated as keypoints with RDP (Fig. \ref{fig:pathcorr}a). After this, a geometric analysis is performed based on the first and last correction keypoints and their closest keypoints of the path model (these are indicated with orange arrows in Fig. \ref{fig:pathcorr}b). A line segment is generated from these closest keypoints to the following model keypoints (cyan dotted line in Fig. \ref{fig:pathcorr}b) and the correction keypoint is projected onto that line segment (blue line in Fig. \ref{fig:pathcorr}b). The redundant keypoints (plotted with red crosses in Fig. \ref{fig:pathcorr}b) are deduced from whether the projection casts onto the line segment. After this, the redundant keypoints are deleted from the path model and the correction keypoints are inserted, producing a new path (dashed blue line in Fig. \ref{fig:pathcorr}c). Algorithm 5 displays further details of this geometrical analysis.

\begin{algorithm}
\caption{Correct a path model's keypoints}\label{alg:pathcorr}
\hspace*{\algorithmicindent} \textbf{Input:} Correction path points, Gaussian path model \\
\hspace*{\algorithmicindent} \textbf{Output:} Adjusted 3-D path model 
\begin{algorithmic}[1]
\State Correction keypoints $\leftarrow$ RDP(Correction path points)
\State Find the closest path model keypoint to the first correction keypoint
\If{closest point is the first point of path model}
    \State Next keypoint $\rightarrow$ \textit{first redundant keypoint}
\ElsIf{closest point is the last point of path model}
    \State Previous keypoint $\rightarrow$ \textit{first redundant keypoint}
\Else{}
    \State Draw line segment between the closest point and the following path model keypoint
    \State Project first correction keypoint onto line segment
    \If{projection is between closest points}
        \State Next keypoint $\rightarrow$ \textit{first redundant keypoint}
    \Else{}
        \State Closest keypoint $\rightarrow$ \textit{first redundant keypoint}
    \EndIf
\EndIf
\State Find the closest path model keypoint to the last correction keypoint
\If{closest point is the first point of path model}
    \State Next keypoint $\rightarrow$ \textit{last redundant keypoint}
\ElsIf{closest point is the last point of path model}
    \State Previous keypoint $\rightarrow$ \textit{last redundant keypoint}
\Else{}
    \State Draw line segment between the closest point and the following path model keypoint
    \State Project first correction keypoint onto line segment
    \If{projection is between closest points}
        \State Closest keypoint $\rightarrow$ \textit{last redundant keypoint}
    \Else{}
        \State Previous keypoint $\rightarrow$ \textit{last redundant keypoint}
    \EndIf
\EndIf
\State \textit{Redundant keypoints} $\leftarrow$ range(\textit{first redundant keypoint}:\textit{last redundant keypoint})
\State Replace the redundant keypoints with the correction keypoints in path model's keypoints
\State \textbf{return} corrected keypoints
\end{algorithmic}
\end{algorithm}

\section{Results}
\label{sec:results}

\subsection{Generation of the path library}\label{result:lib}
\begin{figure*}[htbp]
    \centering
    \includegraphics[width=\textwidth]{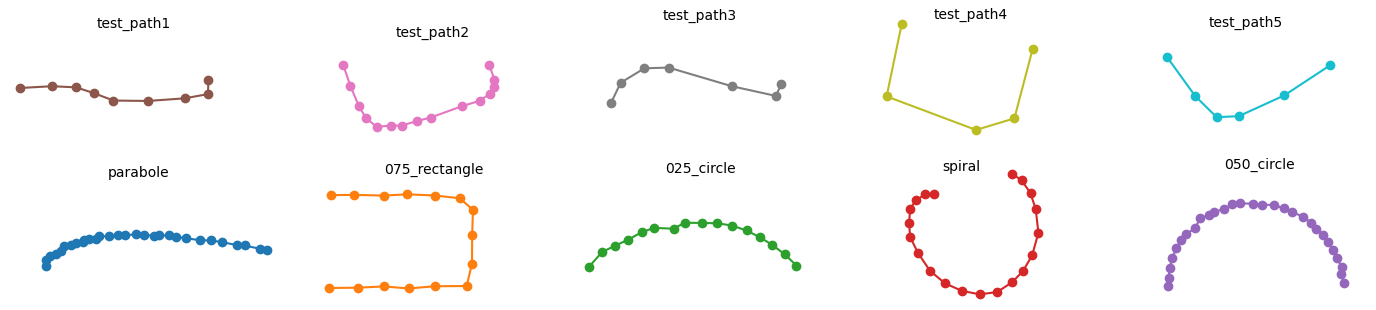}
    \caption{The keypoints of the Gaussian path models in the path recognition library.}
    \label{fig:path_library}
\end{figure*}

A path model library was constructed for path recognition tests consisting of ten path models. Five of the path models were generic geometric shapes (Fig. \ref{fig:path_library}: parabola, rectangle, quarter-circle, spiral, and half-circle) whose path data was synthetic and generated with software. For each of these synthetic paths, the path data consisted of eight to ten teaching sets. Noise was introduced to the data by adding a normally distributed random variable with a mean of 0 to each path point in the data. 

\begin{figure}[htbp]
    \centering
    \includegraphics[width=.8\columnwidth]{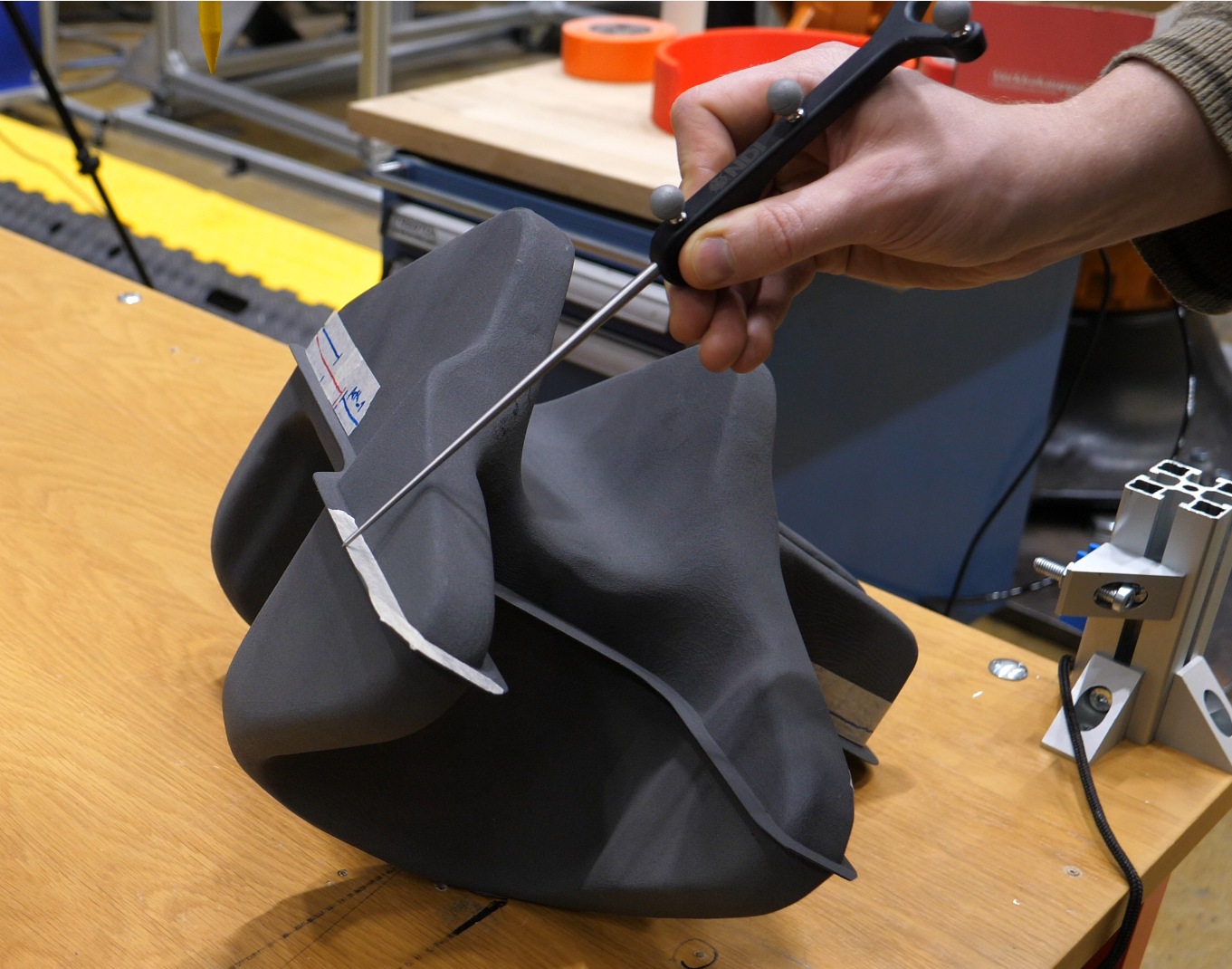}
    \caption{Test object and tracking tool (NDI Polaris Vega XT) used for path model teaching and path demonstrations.}
    \label{fig:testobject}
\end{figure}

The other five path models were taught by human demonstration on a test object using a tracking tool \cite{ndi} (Fig. \ref{fig:testobject}). The shapes of these paths (Fig. \ref{fig:path_library}: test paths 1-5) were arbitrarily selected by choosing sections from edges of the test object and marking them as paths. The tracking tool was used to demonstrate these paths and teaching data was collected during the demonstration. Four to five teaching sets were tracked for each of these paths. Due to the small number of teaching sets, some noise was added to teaching data with a normally distributed random variable. The necessity of having enough variance in the teaching data to create a robust Gaussian model of the path is discussed further in section \ref{disc:teachdata}.
Gaussian path models were generated from the teaching data with Algorithm 2. A suitable level of decimation for the path shapes was found through iteration: the level of decimation was incremented while tracking the score for a correct recognition and the best score of the incorrect path models. By maximizing the difference between these two scores, the level of decimation for each path was found. In addition to maximizing the difference of correct and best incorrect scores, it should be considered that a lower level of decimation provides a geometrically more accurate representation of the original path as it is less approximated. This iterative method for finding the level of decimation is discussed further in section \ref{disc:decim}. 

\subsection{Path demonstrations}\label{result:demos}
Demonstrations corresponding to all of the generated path models in the library consisted of one set of path points. Demonstration data of the five simple geometric shapes was generated with software, much like the teaching data for these paths. Demonstration data of the paths on the test object was collected with human demonstration using the same tracking tool the teaching data was collected with. The test object's pose was different during the collection of path teaching data and during the tracking of the path demonstrations to assess the performance of the path canonicalization. No noise was added to the demonstration data of the paths.

\subsection{Path recognition}\label{result:recog}
\begin{figure}[htbp]
    \includegraphics[width=.7\columnwidth]{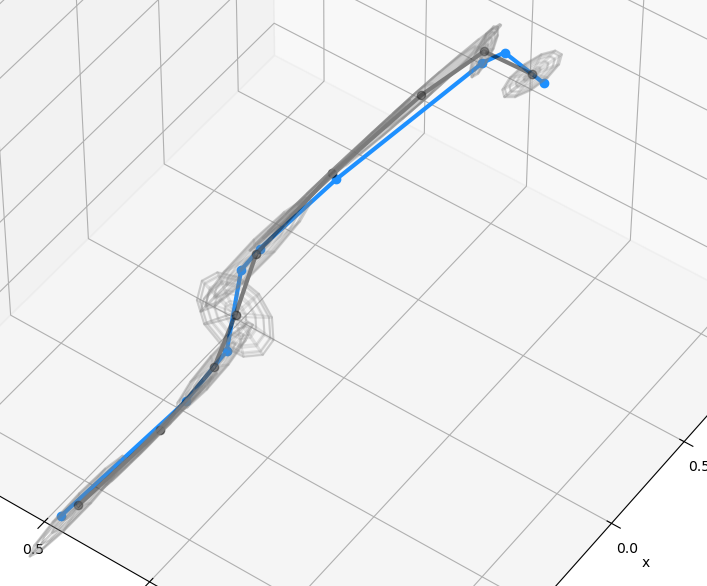}
    \centering
    \caption{A demonstrated path (plotted in blue) recognized as a Gaussian path model from the path recognition library (plotted in black).}
    \label{fig:demoedpath}
\end{figure}

The path recognition algorithm (Algorithm 3) was executed with the demonstration data and the taught path models. Fig. \ref{fig:demoedpath} presents one of the paths demonstrated on the test object in its canonicalized form (plotted in blue), alongside the Gaussian path model from the library it was classified as (plotted in black). The gray ellipsoid shapes represent the covariances of the Gaussian path model's keypoints.

\begin{figure*}[htbp]
    \centering
    \includegraphics[width=.9\textwidth]{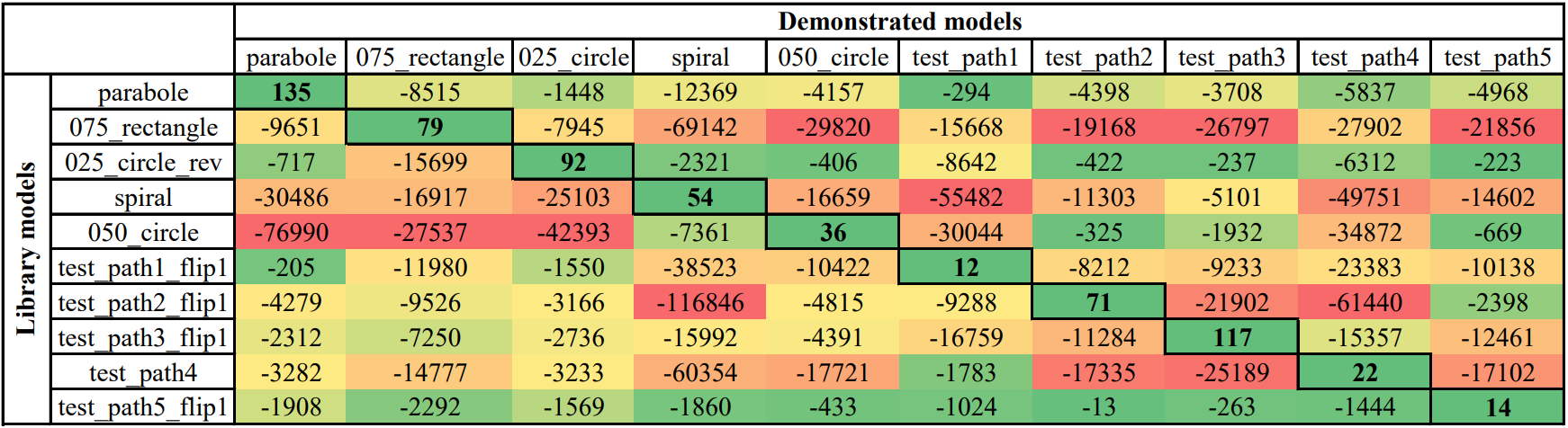}
    \caption{The path recognition algorithm's performance with the selected paths.}
    \label{fig:diagonal}
\end{figure*}

Fig. \ref{fig:diagonal} displays a comparative analysis of the path recognition algorithm’s performance with the selected paths. Each column represents a demonstrated path and each row a path model in the path recognition library. A green-yellow-red color coding signifies the score values for each path classification from good to worse. The best score for each demonstrated path is highlighted in bold. The robustness of the presented path recognition algorithm is emphasized by Fig. \ref{fig:diagonal} as no false classifications were made. 

\subsection{Path correction}\label{result:mod}
\begin{figure}[htbp]
    \centering
    \includegraphics[width=.7\columnwidth]{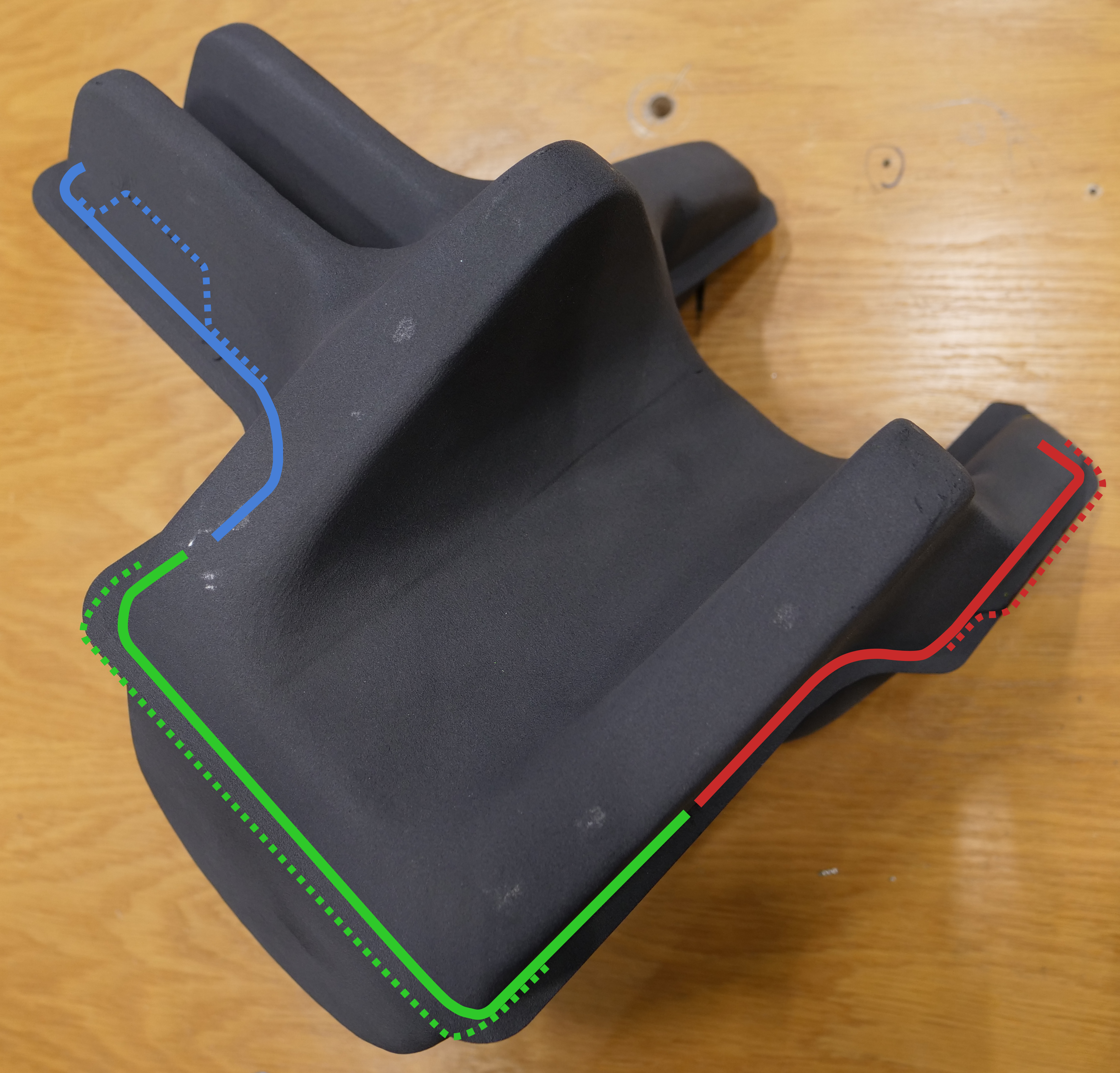}
    \caption{Approximate positions of test path 1 (red), test path 2 (green), and test path 3 (blue) and their respective corrections (dotted lines) on the test object.}
    \label{fig:paths_testobject}
\end{figure}

\begin{figure}[htbp]
    \centering 
    \begin{subfigure}{0.49\columnwidth}
        \includegraphics[width=\textwidth]{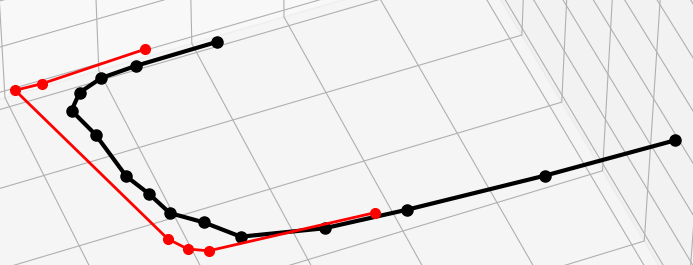}
        \caption{Path model keypoints (black) and correction keypoints (red).}
        \label{fig:path2_democorr}
    \end{subfigure}    
    \hfill
    \begin{subfigure}{0.49\columnwidth}
        \includegraphics[width=\textwidth]{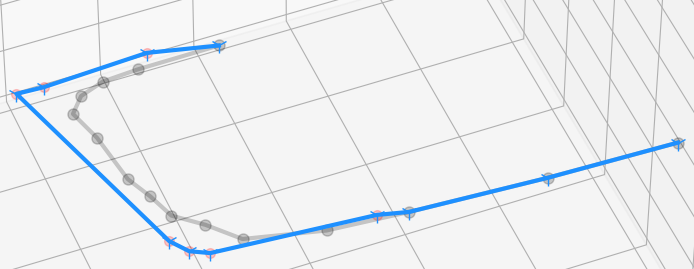}
        \caption{Corrected path's \\keypoints.}
        \label{fig:path2_corrected}
    \end{subfigure}   
    \caption{An example of path correction for test path 2.}
    \label{fig:path2_correction}
\end{figure}

\begin{figure}[htpb]
\sbox\twosubbox{%
  \resizebox{\dimexpr.75\columnwidth-1em}{!}{%
    \includegraphics[height=5cm]{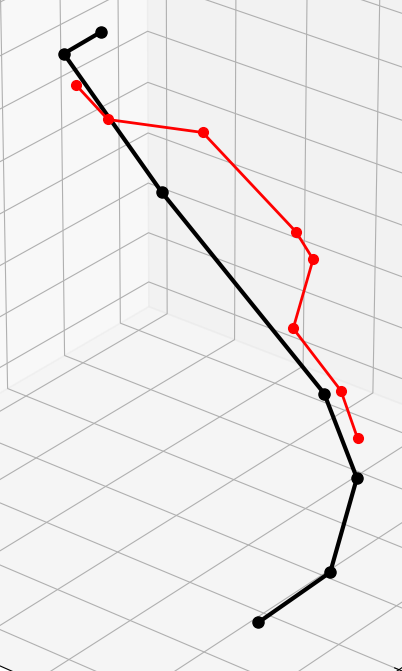}%
    \includegraphics[height=5cm]{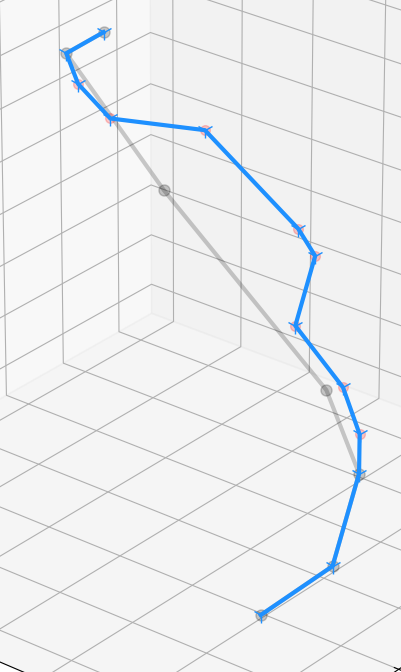}%
  }%
}
\setlength{\twosubht}{\ht\twosubbox}
\centering
\subcaptionbox{Path model keypoints (black) and correction keypoints (red).\label{fig:path3_democorr}}{%
  \includegraphics[height=\twosubht]{path_3_democorr_cropped.png}%
}\quad
\subcaptionbox{Corrected path's keypoints.\label{fig:path3_corrected}}{%
  \includegraphics[height=\twosubht]{path_3_corrected_cropped.png}%
}
\caption{An example of path correction for test path 3.}\label{fig:path3_correction}
\end{figure}

Three paths (test paths 1-3 in Fig. \ref{fig:path_library}) were demonstrated on the test object (Fig. \ref{fig:paths_testobject}) along with corrections to the paths. These paths were classified with the path recognition algorithm (Algorithm 3) and the keypoints of the best-fit model were corrected with the path correction algorithm (Algorithm 5). 
The path model keypoints, the path correction keypoints and the corrected path are displayed in Fig. \ref{fig:pathcorr} for test path 1, Fig. \ref{fig:path2_correction} for test path 2, and in Fig. \ref{fig:path3_correction} for test path 3.

\section{Discussion}\label{sec:discussion}
\subsection{Role of the quality and quantity of teaching data}\label{disc:teachdata}
The method for path recognition requires only a small amount teaching data --- as few as four demonstrations were sufficient for generating a Gaussian path model that was utilizable in path classification. However, for successful path recognition with the generated models, noise was added to the teaching data due to the small amount of teaching data. 
More testing is needed, with larger teaching sets per path, in order to see if more teaching data can substitute the need for adding noise to the data. 
Additionally, a process for investigating and deriving a sufficient level of noise in the teaching data to create a robust Gaussian path model is a matter for further research. For the current tests, a suitable amplitude for the added noise for generating the Gaussian model was found by iteration. However, a systematic method for defining the suitable level of noise in teaching data is necessary for utilizing these methods outside testing and development purposes.

\subsection{Path model decimation}\label{disc:decim}
\begin{figure}[htbp]
    \centering 
    \begin{subfigure}{0.49\columnwidth}
        \includegraphics[width=\textwidth]{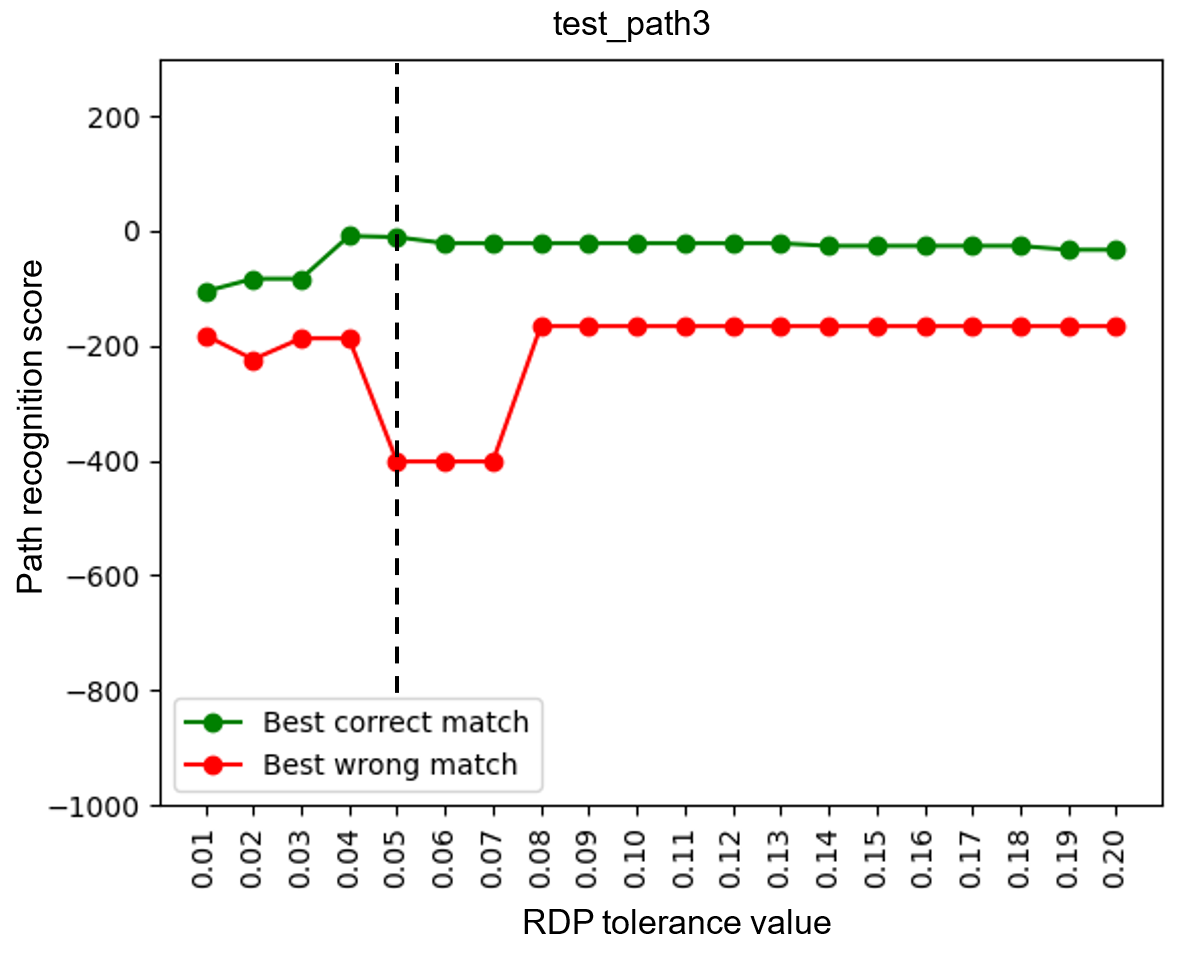}
        \caption{Test path 3.}
        \label{fig:eps_path3}
    \end{subfigure}    
    \hfill
    \begin{subfigure}{0.49\columnwidth}
        \includegraphics[width=\textwidth]{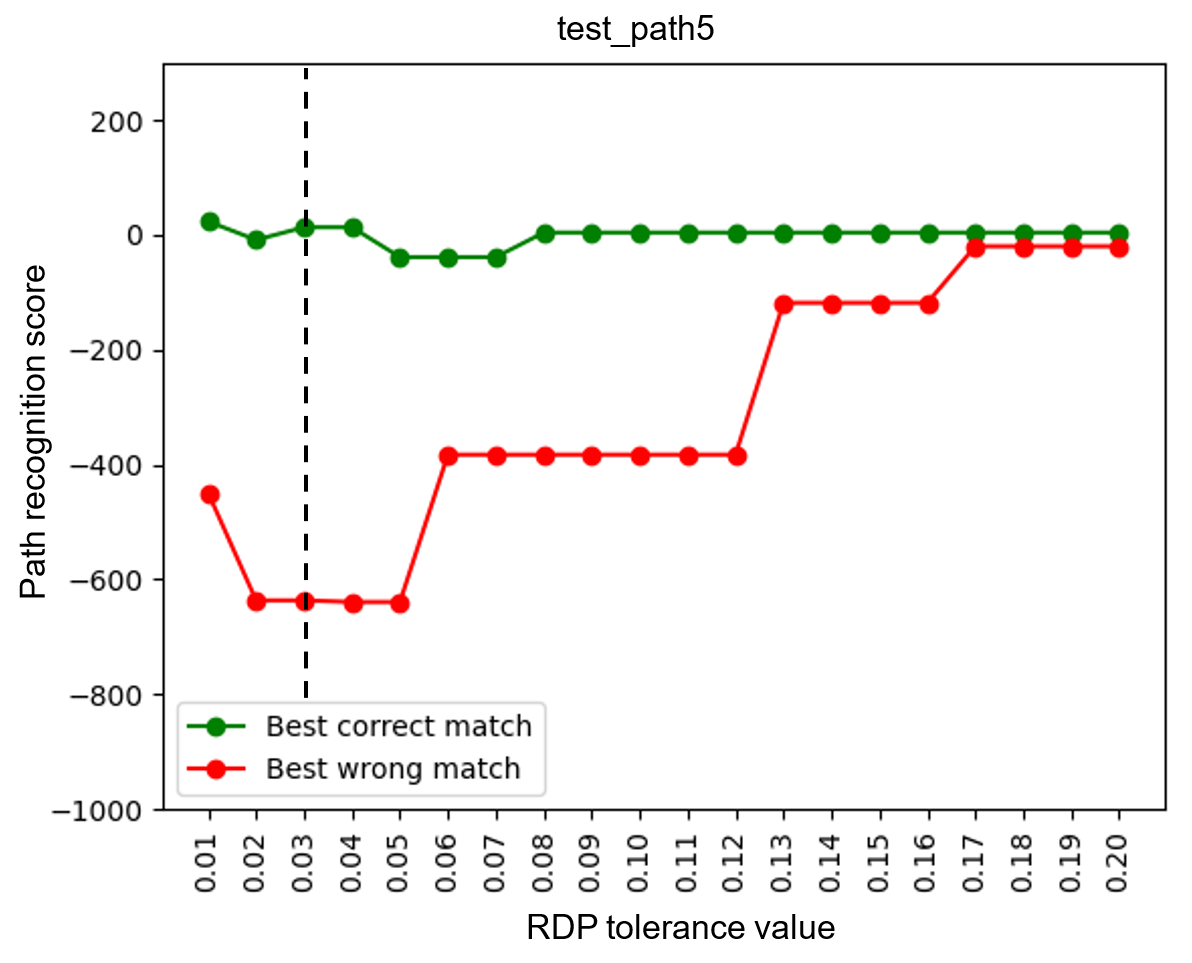}
        \caption{Test path 5.}
        \label{fig:eps_path5}
    \end{subfigure}   
    \caption{The path recognition scores of the correct recognition (green) and the best incorrect recognition (red) as a function of RDP algorithm's tolerance value. The selected RDP tolerance is highlighted with a dashed line.}
    \label{fig:score-epsilon}
\end{figure}

The level of decimation for the path models was found through iteration in the tests by executing the path recognition algorithm while varying the tolerance value of the decimation algorithms and tracking the scores of the correct match and the best-scoring incorrect match (\ref{result:lib}). The difference of these scores was maximized. Fig. \ref{fig:score-epsilon} displays examples of the path recognition score's behaviour as the tolerance value of the RDP decimation algorithm is varied. As the tolerance value is increased, the amount of keypoints in the model is reduced, increasing the score for incorrect recognitions. The selected tolerance value for the decimation is highlighted with a dashed line, where the difference between the correct recognition score and the best score for an incorrect recognition is at its largest. 
It is important to note that this method for finding a suitable level of decimation requires manual work along with the knowledge of a correct match from the path recognition library to the path being demonstrated. Systematic methods for finding the suitable level and decimation for teaching data could be developed. Research on methods that utilized parameters derived from the path's shape (such as path length, angular changes and rate of angular changes) to deduce a tolerance parameter for the decimation algorithms is a possible extension to the work introduced in this paper. 

\subsection{Path modification}
The presented path modification algorithm has a limitation in that the demonstrated correction has to be in the same coordinate frame as the demonstrated path. However, this method allows for utilizing the path library's models in a flexible manner: A path can be demonstrated by hand, and corrected subsequently through providing a new path without the need for teaching the desired path shape. This can be useful for long or complex paths that require small or one-time modifications. Furthermore, in situations where path shape changes (such as grinding, where the object geometry changes with each execution of a path) occur over time, this method may be used to compensate for the geometry changes. 

\subsection{Future Work}
More testing of the path recognition with different paths is needed in order to determine the scope of the required variance in path shapes for generating a path library capable of classifying path demonstrations in most use cases. More investigation into the use of synthetically generated path models for classifying human demonstrations is underway. Preliminary testing on this has already been conducted with the proposed methods. In addition, force values will be introduced in the Gaussian path models to allow the path library to be used for force controlled motion programming by demonstration. Moreover, extending path models with time data and orientations for the path keypoints will be investigated as well. Finally, path segmentation -- using path models in such a library as path primitives -- will be investigated in the future.

\section{Conclusions}\label{sec:conclusions}

Methods for creating Gaussian path models and using those models to classify paths by demonstration were presented in this paper. The methods were validated through generating a path model library from synthetically generated as well as demonstrated path data and by using this library in the classification of generated and demonstrated paths. Results of the testing showed that the presented methods work as intended for both software generated paths and human demonstrations of paths. Additionally, methods for utilizing the Gaussian path models for robot motion programming and modifying existing Gaussian path models were introduced. The objective of this work is that software generated path data (such as path data from 3-D CAD models) and human demonstrations of paths may be used for programming precise robotic motions intuitively with human demonstrations. The results presented are promising and further studies will be conducted.

\section*{Acknowledgment}
This work was supported by VTT Technical Research Centre of Finland and the INVERSE project, funded by the European Union, Horizon Europe research and innovation programme (Grant Agreement 101136067).

\vspace{12pt}

\end{document}